\documentclass{article}
\usepackage{hyperref}
\usepackage{graphicx}
\usepackage{amsmath}
\usepackage{comment}
\usepackage{multirow}
\usepackage{multicol}
\usepackage{setspace}
\usepackage{xcolor}
\usepackage[margin=3.5cm]{geometry}

\DeclareMathOperator*{\argmax}{argmax}
\newcommand{\methodname}{Text2Chart}
\title{\methodname: A Multi-Staged Chart Generator from Natural Language Text}
\author{\small Md. Mahinur Rashid, 
Hasin Kawsar Jahan, 
Annysha Huzzat, 
Riyasaat Ahmed Rahul,\\\small
Tamim Bin Zakir,
Farhana Meem,
Md. Saddam Hossain Mukta and
Swakkhar Shatabda \thanks{\{mrashid171045, hjahan171054, ahuzzat171034, rrahul171089, tzakir171032, fmeem171031\}@bscse.uiu.ac.bd, \{saddam, swakkhar\}@cse.uiu.ac.bd}
\\
\small Department of Computer Science and Enginnering, United International University}
\date{}
\begin{document}

\doublespacing
%
\maketitle              
%
\begin{abstract}
Generation of scientific visualization from analytical natural language text is a challenging task. In this paper, we propose {\methodname}, a multi-staged chart generator method. {\methodname} takes natural language text as input and produce visualization as two-dimensional charts. {\methodname} approaches the problem in three stages. Firstly, it identifies the axis elements of a chart from the given text known as $x$ and $y$ entities. Then it finds a mapping of $x$-entities with its corresponding $y$-entities. Next, it generates a chart type suitable for the given text: bar, line or pie. Combination of these three stages is capable of generating visualization from the given analytical text. We have also constructed a dataset for this problem. Experiments show that {\methodname} achieves best performances with BERT based encodings with LSTM models in the first stage to label $x$ and $y$ entities, Random Forest classifier for the mapping stage and fastText embedding with LSTM for the chart type prediction. In our experiments, all the stages show satisfactory results and effectiveness considering formation of charts from analytical text, achieving  a commendable overall performance.

\end{abstract}
\section{Introduction}
In recent years, advances in Natural Language Processing (NLP) have made huge progress to extract information from natural language texts. Among them a few example tasks are: document summarization \cite{liu2019text}, title or caption generation from texts, generating textual description of charts \cite{balaji}, named entity recognition \cite{sang2003introduction}, etc. There has been several attempts to generate graphs or structural elements from natural language texts or free texts \cite{ghosh2018automated,shaw2019generating,obeid2020chart,guo2020cyclegt}. Scientific charts (bar, line, pie, etc.) are visualizations that are often used in communication. However, automated generation of charts from natural language text always has been a challenging task.


There are very few works in the literature addressing the exact problem of scientific chart generation from natural language text \cite{cui2019text,brown2020language}. In \cite{cui2019text}, the authors have presented a infographics generation technique from natural language statements. However, their method is limited to single entity generation only. {\methodname} extends it to multiple entity generation and thus can generate more complex charts. Nevertheless, Generative Pre-trained Transformer 3 (GPT-3)\cite{brown2020language} has been a recent popular phenomenon in the field of deep learning. OpenAI has designed this third-generation language model that is trained using neural networks. To the best of our knowledge, there has been an attempt to make a simple chart building tool using GPT-3. As its implementation is not accessible yet, the field of information extraction regarding chart creation can be still considered unexplored to some extent. Moreover, the datasets used in GPT-3 is a very large one and the training is too expensive. 


In this paper, we introduce {\methodname}, a multi-staged technique that generates charts from analytical natural language text. {\methodname} works in a combination of three stages.  In the first stage, it recognizes $x$-axis and $y$-axis entities from the input text. In the second stage, it maps $x$-axis entities with its corresponding $y$-axis entities and in the third stage, it predicts the best suited chart type for the particular text input. {\methodname} is limited to three types of charts: bar chart, line chart and pie chart. Tasks in each stage are formulated as supervised learning problems. We have created our own dataset required for the problem. Our dataset is labelled for all three stages of {\methodname}. The dataset is divided into train, validation and test sets. We have used a wide range of evaluation metrics for all the three stages. We have used different combinations of word embeddings like word2vec, fastText, Bidirectional Encoder Representations from Transformers (BERT) with several classifiers or models like Bidirectional Long Short Term Memory (LSTM), Feed Forward Neural Networks, Support Vector Machines and Random Forest. 

The experimental results shows, best results in the first stage are obtained using BERT embedding and Bidirectional LSTM achieving 0.83 of $F1$-score for $x$-entity recognition and 0.97 $F1$-score for $y$-entity recognition in the test set. In the mapping stage, Random Forest achieves best results of 0.917 of Area under Receiver Operating Characteristic Curve (auROC) in the test set. In the third stage, the model fastText with LSTM layers performs the best to predict the suitable chart type. We have observed that bar charts are suitable to all the texts in our dataset. Thus the problem is a multilabel classification problem. However, the second label prediction task is a binary classification task to distinguish between line chart and pie chart. Here, {\methodname} achieves best results of auROC 0.64 for pie charts and auROC 0.91 for line charts.  The experimental analysis of each stage  and in combination shows the overall effective performance of {\methodname} for generating charts from given natural language charts.

The rest of the paper is organized as follows: Section~\ref{secRel} presents a brief literature review of the field and related work; Section~\ref{secMeth} presents the details of the methodology of {\methodname}; Section~\ref{secRes} presents the experimental analysis and discussion on the results and the paper concludes with brief remarks on the limitations and future work in Section~\ref{secCon}. 

\section{Related Work \label{secRel}}
Recent developments in the field of NLP is advancing information extraction in general. One of the first and foremost steps in NLP is the proper vectorization of the input corpora. One of the breakthrough in this area is word2vec proposed in \cite{mikolov2013efficient}. Word2Vec maps words with similar meaning to adjacent points in a vector space. The embedding is learnt using a neural network on continuous bag of words or skip-gram model. A character-level word embedding is proposed in \cite{bojanowski2017enriching}. Recently, Bidirectional Encoder Representations from Transformers (BERT) is proposed in \cite{devlin2018bert}. BERT is trained on a large corpora and enables pre-trained models to be applicable to transfer learning to a vast area of research. BERT has been successfully applied to solve problems like Named Entity Recognition (NER) \cite{sang2003introduction}, text summarization \cite{liu2019text}, etc. 

Text based information processing has been a long quest in the field \cite{kobayashi1999toward}. Kobayashi et el. \cite{kobayashi1999toward} have presented a NLP based modelling for line charts. A Hidden Markov Model based chart (bar, line, etc) recognition method is proposed in \cite{zhou2001learning}. Graph neural networks have been employed in \cite{shaw2019generating} to generate logical forms with entities from free text using BERT. In a very recent work \cite{obeid2020chart}, Obeid et al. have used transformer based models for text generation from charts. For this work, they have also constructed a large dataset extracting charts from Statista. However, their work focuses on chart summarization and hence called `Chart-to-Text'. In an earlier work \cite{huang2007generating}, authors have proposed a method for generating ground truth for chart images. Both of the works are limited to bar charts and line charts only. A Generative Adversarial Network, AttnGAN is proposed in \cite{xu2018attngan} that can generate images from text descriptions. Balaji et al. \cite{balaji} has proposed an automatic chart description generator. 
CycleGT has been proposed recently that works on both directions: text to graphs and graphs to text \cite{guo2020cyclegt}. Kim et al. \cite{kim2020answering} has proposed a pipeline to generate automatic question answering system based on charts.

Automated visualizaton has been always a very fascinating area. A survey of Machine Learning based visualization methods has been presented in \cite{wang2020applying}. Deep Eye is proposed in \cite{luo2018deepeye} to identify best visualizations from pie chart, bar chart, line chart and scatter chart for a given data pattern. A system for automated E-R diagram generation by detecting different entities from natural language text is presented in \cite{ghosh2018automated}. `Text-to-Viz' is proposed in \cite{cui2019text} that generates excellent infographics from given text. However, their method is limited to single entity only. GPT-3 \cite{brown2020language} has been a recent phenomenon in the field which has been reported to generate charts from natural language texts. However, GPT-3 implementation is not open yet. Moreoever, it is trained on an extremely large corpora and an extremely large transformer based model required huge resources. On the light of the review of the existing methods, we believe there is a significant research gap to be addressed in this area.

\section{Our Method \label{secMeth}}
\begin{figure}[!htb]
    \centering
      \includegraphics[width=\textwidth]{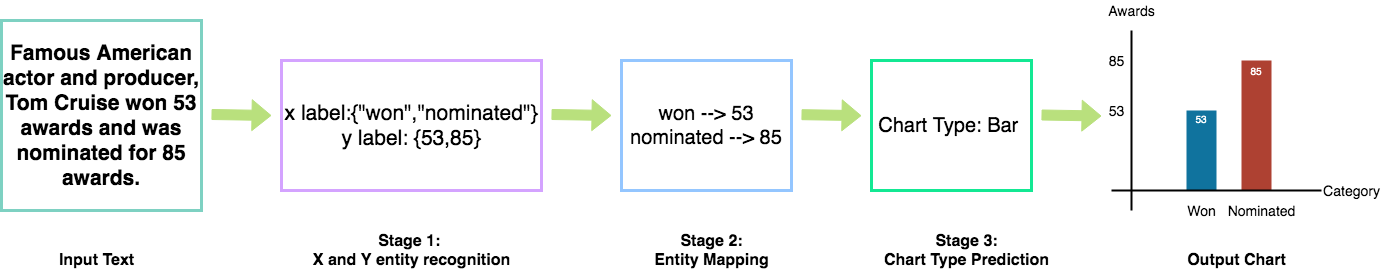}
      \caption{The overall methodology of {\methodname}.}
      \label{archi_1}
    \end{figure}

{\methodname} consists of three stages as shown in Fig.~\ref{archi_1}. It takes a free text as input containing the analytical information. Then it produces $x$ and $y$ axis entities followed by a mapping generation among these elements. In stage 3, the chart type is predicted. A combination of these three are then passed on to the chart generation module. This section presents the detailed procedure of these stages.

\subsection{Stage 1: $x$-Axis and $y$-Axis Label Entity Recognition\label{sec:3.1}}

In the first stage of our technique, we identify the potential candidate words for both $x$-axis and $y$-axis entities of a two dimensional chart. We have formulated the problem as a supervised machine learning task. Here, input to the problem is a paragraph or natural language text and output is a list of words labelled as $x$-entity and $y$-entity. The rest of the words or tokens in the text are ignored. 

To identify $x$-entity and $y$-entity, we build a neural network with different word embeddings and sequence representations. We have employed and experimented with  two different strategies - i) detecting both types of entities at once and ii) using a separate models for recognizing x and y entities. Detecting both $x$ and $y$ entities at once shows a drawback as there lies a possibility that a certain type of entity may outperform the loss function of the other types as observed in the experiments (Section~\ref{secRes1}). 

Since the two types of entity require a different level of skill set, we have observed that the task of recognizing $x$ entity is far more difficult than recognizing $y$ entity and recognizing $x$ entity requires understanding the samples more deeply than it is required for $y$ entity recognition. Moreover, the sample space of $x$ entity is much larger than $y$ entity. Therefore, we use separate models for recognizing $x$ and $y$ entities. This later approach outperforms the performance of the former one as we can see in the result section (Section~\ref{secRes1}).

\begin{figure}[!htb]
    \centering
      \includegraphics[width=0.45\textwidth]{ 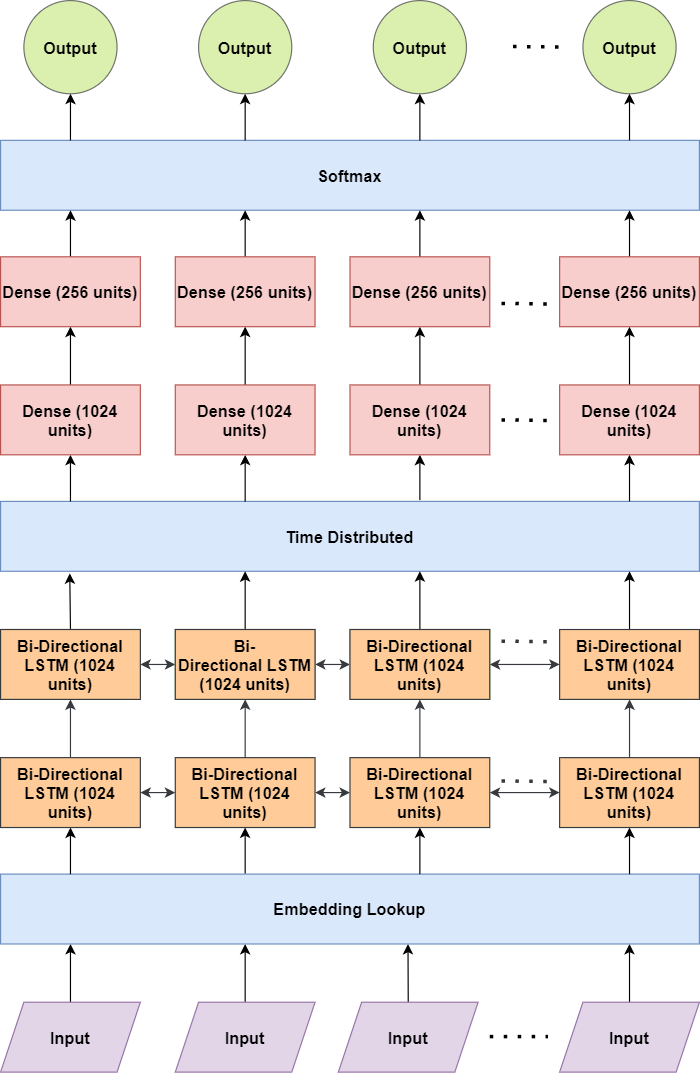}
      \caption{Proposed Neural Architecture for Recognition of $x$-axis and $y$-axis Entities.}
      \label{archi}
    \end{figure}

We have experimented both of the strategies using word embedding like  Word2Vec \cite{mikolov2013efficient}, fastText \cite{bojanowski2017enriching} and the sequence output of the pre-trained model provided by BERT \cite{devlin2018bert}. For each sample text in the dataset, we take the generated embedding  and use it as an input to our model. Then we use layers of Bi-directional LSTM networks. On top of that, we use the time-distribution layer and dense layer to classify each word index that falls into a category of a respected entity or not. The proposed architecture for the first stage of {\methodname} is shown in Fig.~\ref{archi}. Note that the last layer of softmax labels each token either as $x$-entity, $y$-entity or none. 

\subsection{Stage 2: Mapping of $x$ and $y$ Label Entities}
After identifying the $x$ and $y$ entities in Stage 1, we map each of the identified $x$ entity with its corresponding $y$ entity. While inspecting the data samples we build our first intuition that $x$ entities and $y$ entities may not appear in a text sample sequentially as they are mapped but independent of their entity type. 

For example, if we have an $x$ entity set for a text as $\{x_1, x_2, \cdots, x_M\}$ and $y$ entity set of that text is $\{y_1, y_2, \cdots, y_N\}$ and their mapping is as follows $\{(x_1, \phi(x_1)), (x_2, \phi(x_2)), \cdots, (x_M, \phi(x_M)\}$. Please note, here $x_i,y_j$ denotes their position in the sequence. Here the mapping function, $\phi(x_i)$ maps an entity $x_i$ to another entity, $y_k$. However, there is often found that the entity set lengths are not same $M\neq N$ and often the sequential order is not maintained. For two $x$ entities $x_i, x_j$ if they maps to $y_k,y_l$, then a sequential mapping $\phi$ guarantees, $i\le j,  k\le l$ whereas the non-sequential mapping will not guarantee that. However, in our observation, non-sequential mapping is not that frequent. In order to address these issues, we propose that the mapping is dependent on the distances between the corresponding entities. We call it our baseline model for this task. From the training dataset, we learn the probability distribution for positive and negative likelihood for distances between $x$ and $y$ entities which are $P(d(x_i,y_k)|\phi(x_i)=y_k)$ and $P(d(x_i,y_k)|\phi(x_i)\neq y_k)$ respectively. For the missing values in the range, nearest neighbor smoothing is used to estimate the likelihood values and then normalized to convert it to a probability distribution. The baseline model defines the mapping as in the following equation:
\begin{equation}
    \phi(x_i) = \argmax_k \frac{P(d(x_i,y_k)|\phi(x_i)=y_k)}{P(d(x_i,y_k)|\phi(x_i)=y_k)+P(d(x_i,y_k)|\phi(x_i)\neq y_k)}
    \label{eqmax}
\end{equation}



With the initial encouraging results from this simple baseline model (results are shown in Section~\ref{secRes2}), we further extend this and formulate the problem as a supervised learning problem. Now each pair of entities, $(x_i,y_k)$ are converted to a feature vector suitable for supervised learning setting to find that if that pair is mapped or not. 

For a particular entity $x_i$ and a particular $y_k$ entity, we take the two other entities , one immediate before $(x_{i-1}, y_{k-1})$ and the next one $(x_{i+1}, y_{k+1})$ to create the feature vector. For 6 such entity positions, we generate 15 possible pairs and take pairwise distances among them. Note that, for two similar type entities we take unsigned distance and for different entities signed distances are taken to encode their relative positions into the feature vector. With this feature vector, we train two models: SVM and Random Forests, where the latter works slightly better. As this is an argmax based calculation, the probability distribution of Random Forest classifier was more consistent than that of SVM. The reason of the inconsistency of the distribution with the scores in SVC is that,  the `argmax' of the scores may not be the argmax of the probabilities. Therefore we take the auROC as the primary evaluation metrix for this stage. we take the harmonic mean of auROC of both training and validation so that the measure is balanced and they do not outperform each other.







\subsection{Stage 3: Chart Type Prediction}
While generating a chart, we should be aware that type of a chart depends on the information that is conveyed, and the way it is conveyed. Therefore, this sub-task is defined to predict the seemingly appropriate chart type from a text among the most common ones: bar chart, pie chart, and line chart.

\begin{figure}[hbt!]
    \centering
      \includegraphics[width=0.8\textwidth]{ 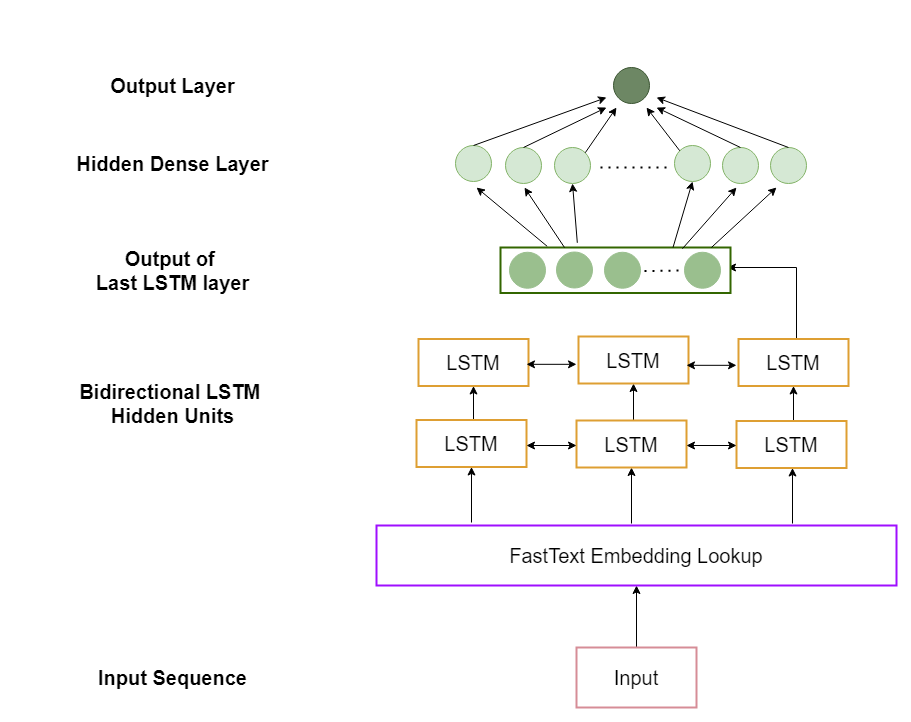}
      \caption{Proposed Neural Network Architecture for Chart Type Prediction.}
      \label{archi_stage3}
    \end{figure}

Generally, a bar chart is the most commonly accepted chart type for any statistical data. However, for better visualization and understanding, pie charts and line charts are also used. Pie charts are suitable if the entities conform to a collection / composition. Line charts are suitable for the cases where the entities are themselves form a continuous domain. For this stage, we have applied fastText word embeddings to build two models with LSTM layers and dense layers. Each model performs binary classification; one is to predict if a pie chart is suited for the text or not, and the other is for line chart. When neither of these two chart types are fitting, only bar chart is assigned to the text. 
The proposed architecture for stage 3 is shown in Fig.~\ref{archi_stage3}. The details of the architecture and experiments are given in Section~\ref{secRes3}.

\section{Experimental Analysis\label{secRes}}
{\methodname} is implemented using Tensorflow version 2.3. All the experiments have run using Google Colab and the cloud GPU provided with it.  The hardware environment of our work has required a CPU of 2.3 GHz, GPU 12 GB, RAM 12.72 GB and Disk of 107 GB. All the experiments have run at least 5 times with different random seeds and only the average results are reported in this section. Source codes and the dataset of {\methodname} will be made available via a public repository (at the time of publication). In the rest of this section, first we describe the process of dataset construction in Section~\ref{secData}, then the performance evaluation methods and metrics are presented in Section~\ref{secEval}. The detailed experimental results of the three stages and overall performance analysis are next presented in respective sections.

\subsection{Dataset Construction\label{secData}}
When we have started this work, no datasets were available for this particular task of automatic generation of a chart out of a natural language text. {\methodname} requires a specific dataset from which the text samples are suitable for recognizing the chart information. Here chart information refers to the $x$-axis entities and the corresponding $y$-axis values respectively. The text samples must contain all these entities to construct the particular chart. We have collected text samples from Wikipedia, other statistical websites and crowd sourcing. We have used crowd sourcing to label the data so that the texts are labelled for all three stages. All the labelled data are then crosschecked by a team of volunteers and only the consensus labels are taken. In total, 717 text samples are taken in the final dataset with 30,027 words/tokens. The average length of the text samples is 53 words and the maximum length is 303 words in a single text. This final dataset is then split in the train, validation and test sets each containing 464, 116 and 137  samples respectively. A summary of the dataset is shown in Table~\ref{instances}. Please note that in the first stage the token number is higher than labels since a particular $x$ or $y$ entity/label might consists of two words or tokens. All the texts are labelled to be suitable for bar charts and only the statistics for pie and line charts are shown in the table.

\begin{table}[t]
\centering
\caption{Summary of datasets used in the experiments.\label{instances}}
\begin{tabular}{|p{1.5cm}|p{1.3cm}|p{1.3cm}|p{1.3cm}|p{1.3cm}|p{1.3cm}|p{1.4cm}|p{0.6cm}|p{0.6cm}|}
    \hline
    &\bf text&\multicolumn{4}{c|}{\bf $x,y$ entity prediction}& \bf mapping&\multicolumn{2}{c|}{\bf chart type}\\ \cline{3-6}
    \cline{8-9}
    \bf dataset&\bf samples&$x$ tokens&$y$ tokens&$x$ labels &$y$ labels& \bf pairs&pie&line\\ \hline
    Training &464&3411	&3614	&1984	&1909	&1984	&73	&58\\
    Validation&116&985	&1058	&548	&529	&548	&20	&11\\
    Test&137&988	&1075	&574	&561	&574	&20	&15\\
    \hline
\end{tabular}
\end{table}

\subsection{Performance Evaluation\label{secEval}}
As {\methodname} is multi-staged and the tasks and related datasets used in the stages are different in nature, they require several different evaluation metrics suitable for particular stage/task in order to evaluate the performance properly. All the methods are trained using the training set and the performance are validated using the validation set. Only after the final model is selected, the model is tested on the test set. 

For axis entity recognition task in the first stage, we adopt the $F1$-score and its variant the harmonic mean of f1-scores. We observe the Receiver Operating Characteristic (ROC) curve and the area under curve (auROC) in order to summarize and compare the performances of the  classifiers in the second stage of entity mapping. Finally for chart type prediction, we adopt Matthews Correlation Coefficient (MCC) evaluation metric, as MCC being a more reliable statistical rate than F1-score and accuracy in binary classification evaluation for imbalanced dataset \cite{chicco2020advantages}.

\subsection{Axis Label Recognition Task\label{secRes1}}

The first stage of our work is $x$-axis and $y$-axis label entity recognition. Here we predict whether a given word from the text input can be an $x$-axis or $y$-axis entity. We have experimented with our neural architecture model of bidirectional LSTM combining several embeddings, such as fastText, Word2Vec and BERT in order to recognize these entities. For each of the embedding, we have used two different approaches. In the first approach, $x$-entity and $y$ entity prediction is considered as separate prediction tasks. Here we have the two models, one for each of the tasks. In the second approach, they are considered together as a combined prediction task. 

\subsubsection{Experiments with fastText embedding}
For both of the approaches using fastText (individual and combined), we have used a neural architecture with 4 hidden layers and a dense output layer. The first two hidden layers consist of bidirectional LSTM layers of 512 neurons and 128 neurons followed by time distributed dense layer of 64 neurons and a dense hidden layer with 1024 neurons. Epoch and batch size are kept fixed at 8 for all the models considered here. Experimental results of fastText experiments are given in the first four rows of Table~\ref{tabStage1}. Note that we have reported precision, recall and $F1$-score for $x$ and $y$ entity predictions. Also the harmonic mean of $F1$-score is reported. Note that, the individual approach achieves $F1$-score for $x$ and $y$ entities of 0.66 and 0.85 respectively in the validation set which is improved in the combined approach being 0.66 and 0.89. It is clear that the prediction or recognition of $x$ axis entities is much difficult task compared to $y$ axis entity recognition. Here, we can conclude that both models perform almost similar which is also reflected in the harmonic mean of $F1$-score respectively 0.74 and 0.76.

\begin{table}[!htb]
    \centering
    \caption{Experimental results for the axis label prediction task in the frist stage of {\methodname}.}
    \label{tabStage1}
\begin{tabular}{|p{1.5cm}|p{1.5cm}|p{1.3cm}|p{1cm}|p{1.3cm}|p{1cm}|p{0.9cm}|p{0.9cm}|p{1.4cm}|}
\hline
    model&dataset&Precision ($x$)&Recall ($x$)&Precision ($y$)&Recall ($y$) &$F1$-score ($x$)&$F1$-score ($y$)&Harmonic  $F1$-score \\
\hline
\hline 
fastText & training &0.81&0.80&0.93&0.88&0.80&0.90&0.84\\
\cline{2-9}
individual& validation 
&0.68&0.64&0.89&0.81&0.66&0.85&0.74\\
\hline 
fastText & training 
&0.81&0.73&0.89&0.97&0.77&0.93&0.84\\
\cline{2-9}
combined& validation 
&0.73&0.60&0.86&0.93&0.66&0.89&0.76\\
\hline 
\hline 
word2Vec & training 
&0.90&0.88&1.00&1.00&0.89&1.00&0.94\\
\cline{2-9}
individual& validation 
&0.72&0.62&0.79&0.77&0.67&0.78&0.72\\
\hline 
word2Vec & training 
&0.99&0.99&1.00&1.00&0.99&1.00&0.99\\
\cline{2-9}
combined& validation
&0.72&0.64&0.83&0.74&0.68&0.78&0.73\\
\hline 
\hline 
BERT & training 
&0.99&0.99&.99&0.99&0.99&0.99&0.99\\
\cline{2-9}
individual& validation 
&\bf 0.89&\bf 0.86&0.95&\bf 0.98&\bf 0.87&\bf 0.97&\bf 0.92\\
\hline 
BERT & training 
&0.99&1.00&0.99&1.00&0.99&0.99&0.99\\
\cline{2-9}
combined& validation 
&0.86&0.78&\bf 0.96&0.97&0.82&\bf 0.97&0.89\\
\hline \hline 
\bf best& \bf test & 0.85&0.82&0.96&0.98&0.84&0.97&0.89\\
\hline 
\end{tabular}
\end{table}

\subsubsection{Experiments with word2vec embedding}

The word2vec embedding represents the word tokens in the corpus by representing the words with common context in a close proximity in the vector space as well. Similar to the experiments of fastText we have two approaches employed here: individual and combined. For word2vec embedding, the network structure is kept the same as in the fastText experiments. However, for training we have used 16 epochs and batch size of 8. The experimental results are shown in the second four rows of Table~\ref{tabStage1}. From Table~\ref{tabStage1}, we can see that here combined approach is giving $F1$-score of $x$ and $y$ entity recognition task as 0.68 and 0.78 respectively which is almost similar to the performance of the individual approach (0.67 and 0.78 respectively). The performance only differ in the $x$ entity recognition task which is also observed in the harmonic mean of $F1$-score. Note that the overall performance of word2vec embedding is significantly worse compared to fastText embedding. Also note that the higher level of overfitting of the word2vec model has reflected in the high values of precision, recall and $F1$ score in all the tasks in the training dataset which is not repeated in validation.   

\subsubsection{Experiments with BERT embedding}

We have also experimented with BERT embeddings on the same architecture proposed in Section~\ref{secMeth}. However, in these experimetns the network structure is different with same number of layers. Here too we have used two approaches: individual and combined. In the individual approach, the first two hidden layers of the neural architecture are bidirectional LSTM with 1024 neurons in each followed by a time distributed dense layer with 1024 neurons and a dense layer with 256 neurons. In the case of $x$ entity recognition, we have used a batch size of 2 and 80 epochs for training. In the case of $y$ entity recognition, the batch size was 8. In the combined approach, the architecture structure has differed only in the last hidden dense layer. Here the number of neurons were 1024. We have used an online training for this combined approach. The experimental results with BERT embedding is reported in the third four rows of Table~\ref{tabStage1}. From the results shown there, we can notice that for BERT embedding, the performances in the individual approach outperform the combined approach in $x$ entity prediction performance. The results in $y$ entity recognition is almost similar for both of the approaches. Thus the both harmonic mean and $F1$-score of $x$ entity recognition are superior in combined approach which are 0.87 and 0.92 respectively compared to those of 0.82 and 0.89 in the individual approach.

To summarize, we can note that the results in BERT embedding are superior to two other embeddings. The best achieved values are shown in bold faced fonts in the Table. Thus, we take the BERT embedding individual $x$ and $y$ entity prediction approach with bidirectional LSTM as the best performing model among those used in the experiments. With the best model, we have also tested its performance on the test dataset. The results are shown in the last row of Table~\ref{tabStage1}. Here, it is interesting to note that the learned model is not overfitting and the performances in the validation set and test set are not much differing.  

\begin{figure}[!hbt]
    \centering
      \includegraphics[width=0.7\textwidth]{ 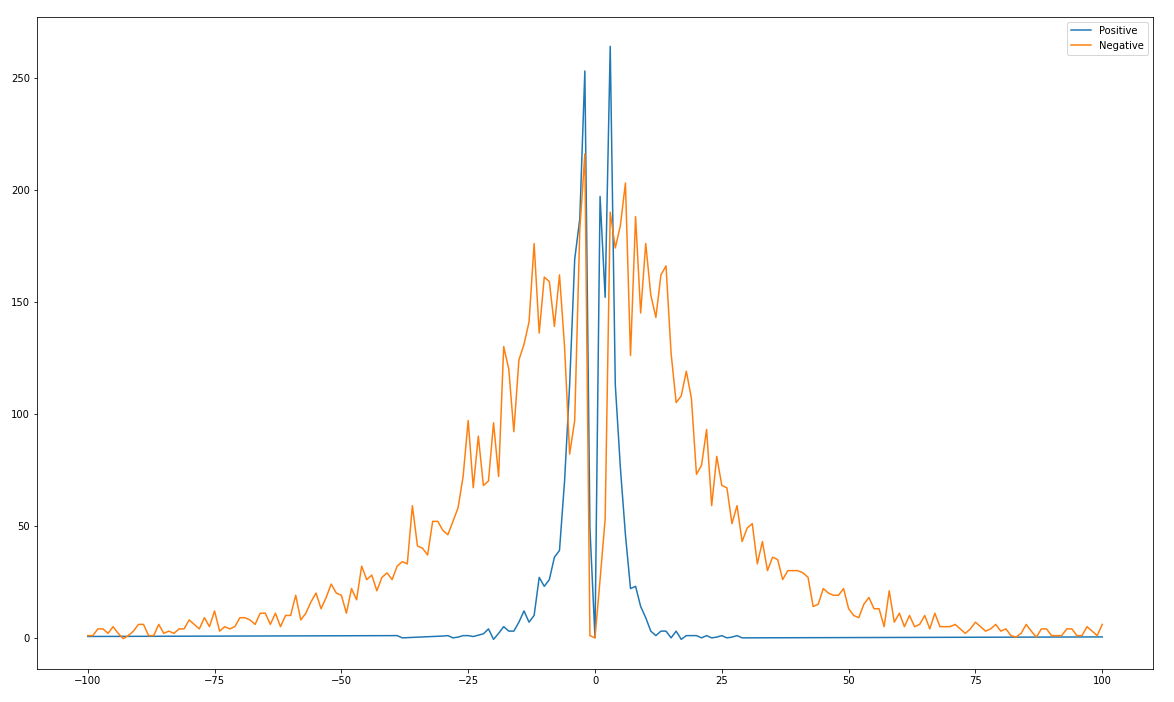}
      \caption{Distribution of positive and negative likelihood frequencies of the entity pairs over their distances, here positive and negative distances denote the sequential order of positions of the entities in the text.}
      \label{data_dis}
    \end{figure}

\subsection{Mapping Task\label{secRes2}}

After recognizing the $x$ and $y$ entities with high precision and recall in stage 1, the second stage sets the target to map them in an ordered way. We have first used a transfer model from the best performing model in first stage to see if that helps. However, the very low $F1$-score of 0.41 and auROC of 0.64 have discouraged to proceed further in this way. It is evident that the same architecture is not suitable for the  different stages due to difference in the type of the task. Note that, this task is highly imbalanced as the number of positive mappings are very small compared to negative mappings. Thus the model often gets biased towards the negative model and might show poor performance in the positive prediction. 

\begin{table}[!htb]
\centering
\caption{Experimental results for the mapping task in the second stage.}
\label{tabresStage2}
\begin{tabular}{|c|c|c|c|c|c|p{1.8cm}|p{1.5cm}|}
    \hline
    \bf model&\bf dataset&\bf class &\bf Precision &\bf Recall &\bf $F1$-score&\bf Harmonic $F1$-score&\bf auROC \\ \hline \hline
    baseline&\multicolumn{1}{c|}{\multirow{2}{*}{training}}&0 (-ve)&0.94&0.94&0.94&{\multirow{2}{*}{0.84}}&{\multirow{2}{*}{0.908}}\\
    &\multicolumn{1}{c|}{}&1 (+ve)&0.76&0.76&0.76&&\\ \cline{2-8}
    &\multicolumn{1}{c|}{\multirow{2}{*}{validation}}&0 (-ve)&0.95&0.95&0.95&{\multirow{2}{*}{0.82}}&{\multirow{2}{*}{0.914}}\\
    &\multicolumn{1}{c|}{}&1 (+ve)&0.73&0.73&0.73&&\\ \hline \hline
    SVM&\multicolumn{1}{c|}{\multirow{2}{*}{training}}&0 (-ve)&0.93&0.93&0.93&{\multirow{2}{*}{0.81}}&{\multirow{2}{*}{0.897}}\\
    &\multicolumn{1}{c|}{}&1 (+ve)&0.72&0.72&0.72&&\\ \cline{2-8}
    &\multicolumn{1}{c|}{\multirow{2}{*}{validation}}&0 (-ve)&\bf 0.96&\bf 0.96&\bf 0.96&{\multirow{2}{*}{\bf 0.86}}&{\multirow{2}{*}{0.924}}\\
    &\multicolumn{1}{c|}{}&1 (+ve)&\bf 0.78&\bf 0.78&\bf 0.78&&\\ \hline 
    \hline
    
    
    Random&\multicolumn{1}{c|}{\multirow{2}{*}{training}}&0 (-ve)&0.95&0.95&0.95&{\multirow{2}{*}{0.85}}&{\multirow{2}{*}{0.913}}\\
    Forest&\multicolumn{1}{c|}{}&1 (+ve)&0.77&0.77&0.77&&\\ \cline{2-8}
    &\multicolumn{1}{c|}{\multirow{2}{*}{validation}}&0 (-ve)&\bf 0.96&\bf 0.96&\bf 0.96&{\multirow{2}{*}{0.84}}&{\multirow{2}{*}{\bf 0.930}}\\
    &\multicolumn{1}{c|}{}&1 (+ve)&0.77&0.77&0.77&&\\ \hline
    \hline 
\bf best & \multicolumn{1}{c|}{\multirow{2}{*}{\bf test}}&0 (-ve)&0.94&0.94&0.94&{\multirow{2}{*}{0.85}}&{\multirow{2}{*}{0.917}}\\
    &\multicolumn{1}{c|}{}&1 (+ve)&0.77&0.78&0.77&&\\
\hline 
\end{tabular}

\end{table}

The baseline model that we try here is based on the probability distribution of the positive and negative likelihoods of the mapped entities Fig.~\ref{data_dis}. Note that, there is an overlapping area among positive and negative occurrences over the distances. Also note that most of the mappings are in relatively short distances or proximities. Based on that our baseline model is a simple argmax calculation of the likelihood based on Eq.~(\ref{eqmax}). The results of the baseline model is presented in the first four rows of Table~\ref{tabresStage2}. In this table, we have reported precision, recall and $F1$-score for both of the classes and also the auROC. Note that the results of the baseline model is encouraging with a high auROC of 0.908. However, note that the positive class performance is poor compared to the negative class which leaves room for improvement.

Next we have experimented with the supvervised learning approach describe in Section~\ref{secMeth} using Support Vector Machine (SVM) and Random Forest classifiers. In Table~\ref{tabresStage2}, we notice that the performance in both of the classes are improved using this approach in both of the classes compared to the baseline model. We note that the performance in the negative class is same. However, $F1$-score of the Random Forest classifier is slightly lower in case of positive case which is not that significant (0.77 vs 0.78). The fact is evident in auROC. There we see significant improvement achieved by Random Forest classifier compared to SVM. The best values are shown in bold faced font in the table. Thus we conclude that Random Forest is the best performing model for stage 2.

\begin{figure}[t]
    \centering
    \begin{tabular}{ccc}
\includegraphics[width=0.27\textwidth]{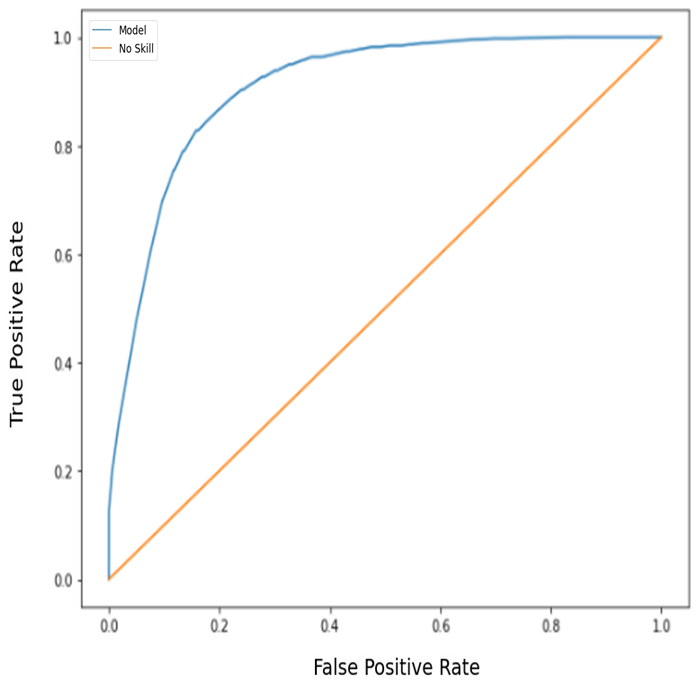}&
\includegraphics[width=0.35\textwidth]{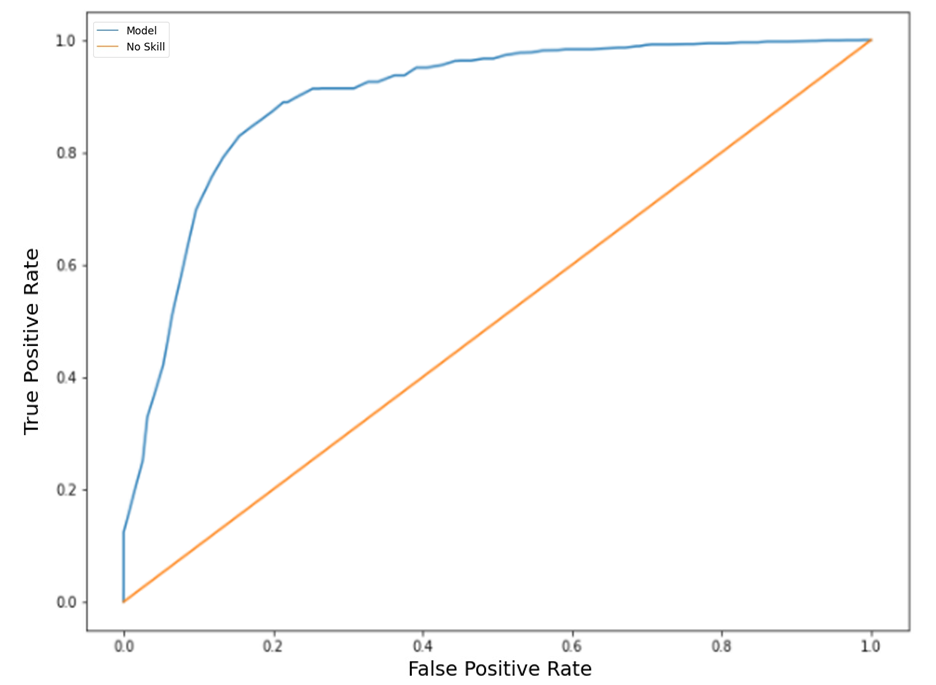}&
\includegraphics[width=0.29\textwidth]{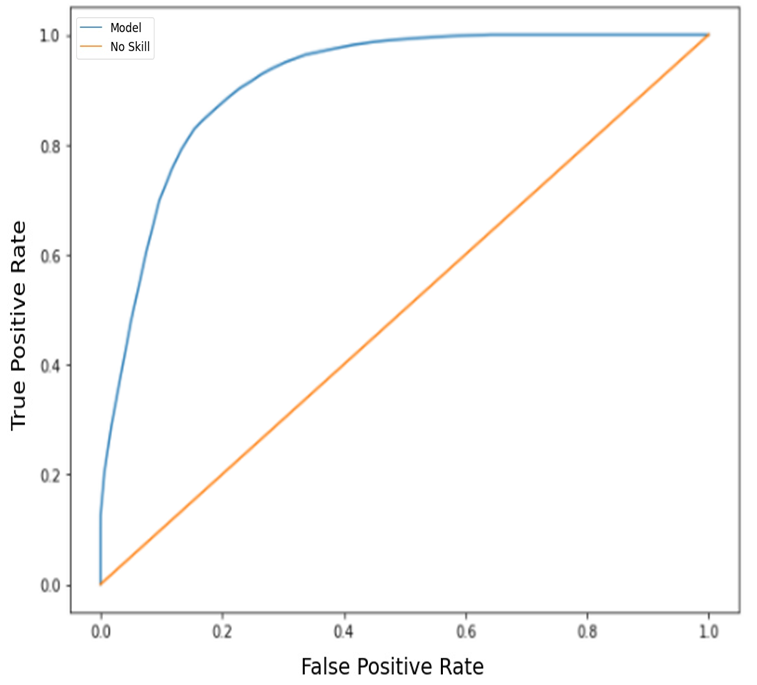}\\
(a)&(b)&(c)\\

\includegraphics[width=0.27\textwidth]{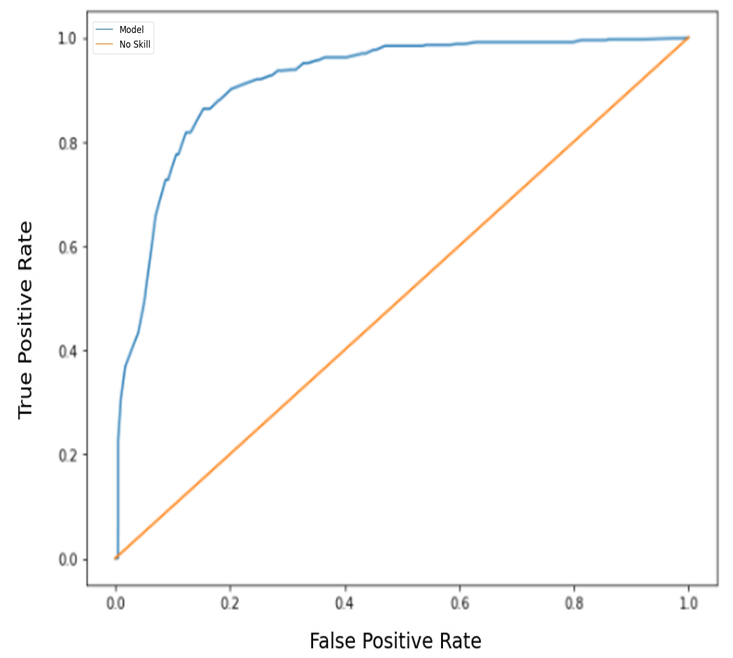}&
\includegraphics[width=0.31\textwidth]{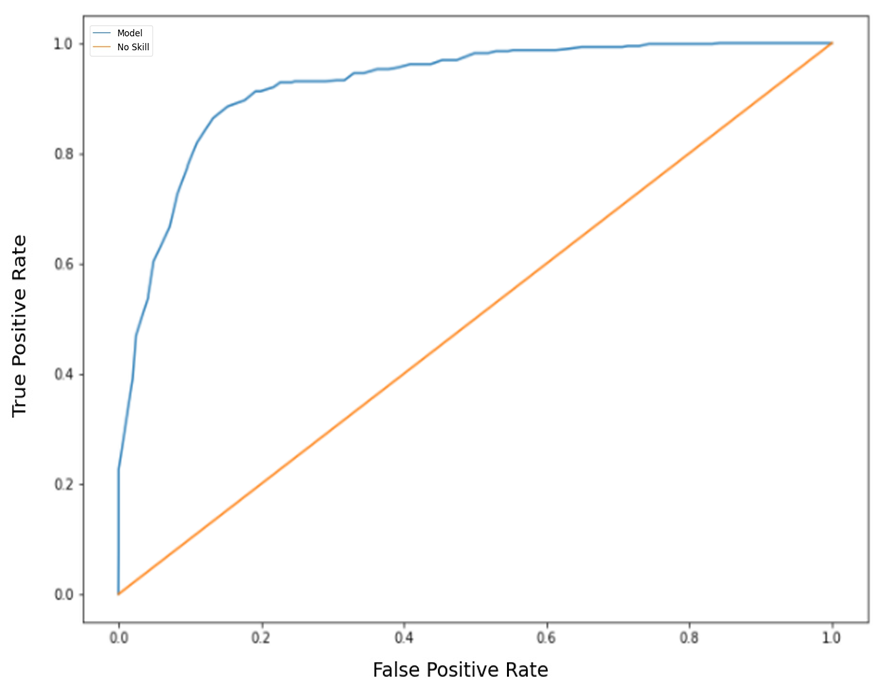}&
\includegraphics[width=0.27\textwidth]{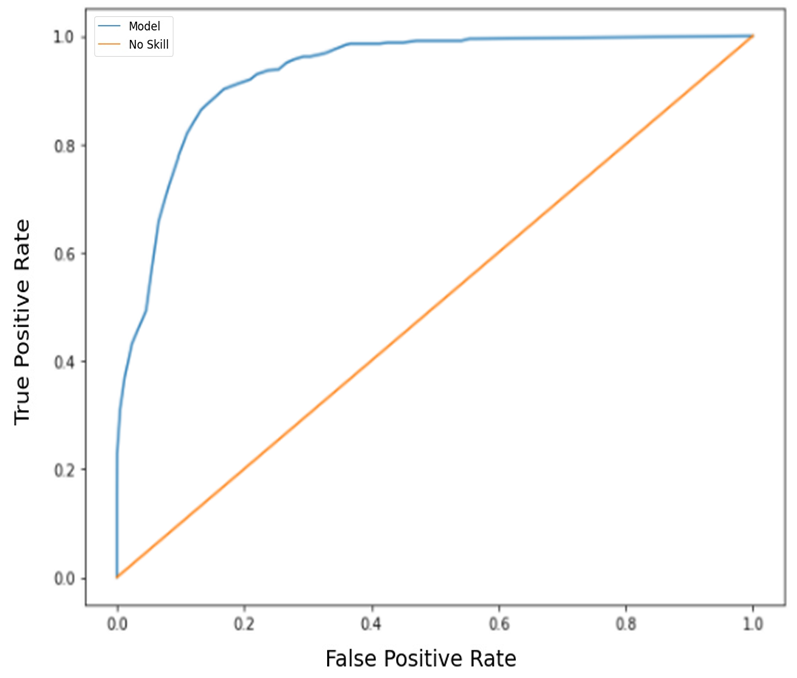}\\
(d)&(e)&(f)\\
    \end{tabular}
    \caption{Receiver Operating Characteristic Curves for training set performances of (a) Probability Model, (b) Support Vector Machines, (c) Random Forest and for validation set of (d) Probability Model, (e) Support Vector Machines, and (f) Random Forest.}
    \label{fig:ROCstage2}
\end{figure}

Finally, we have tested the performance of the best performing Random Forest model on the test set and the results are shown in the last row of the Table~\ref{tabresStage2}. We see that the performances in the test set are stable and similar to validation set. The ROC curves for training and validation set on all models are given in Fig.~\ref{fig:ROCstage2}.

\subsection{Chart Type Prediction Task\label{secRes3}}
At the third stage, the task is to predict the suitable chart type from the given text. Note that for all the texts in the dataset, bar chart is common and thus we exclude it from classification models. We train two separate models: one for the pie chart and another for the line chart. The architecture of the model is shown in Fig. \ref{archi_stage3}. This model uses fastText embedding with bidirectional LSTM layers. The network architecture and structure is kept same for both of the classifiers. The neural network has three hidden layers. The first two layers are the LSTM layers with 128 neurons each followed by a dense layer of 512 neurons. The output layer is a simple sigmoid layer. We have used RMSprop algorithm to train the models. 

\begin{figure}[!htb]
    \centering
    \begin{tabular}{ccc}
\includegraphics[width=0.33\textwidth]{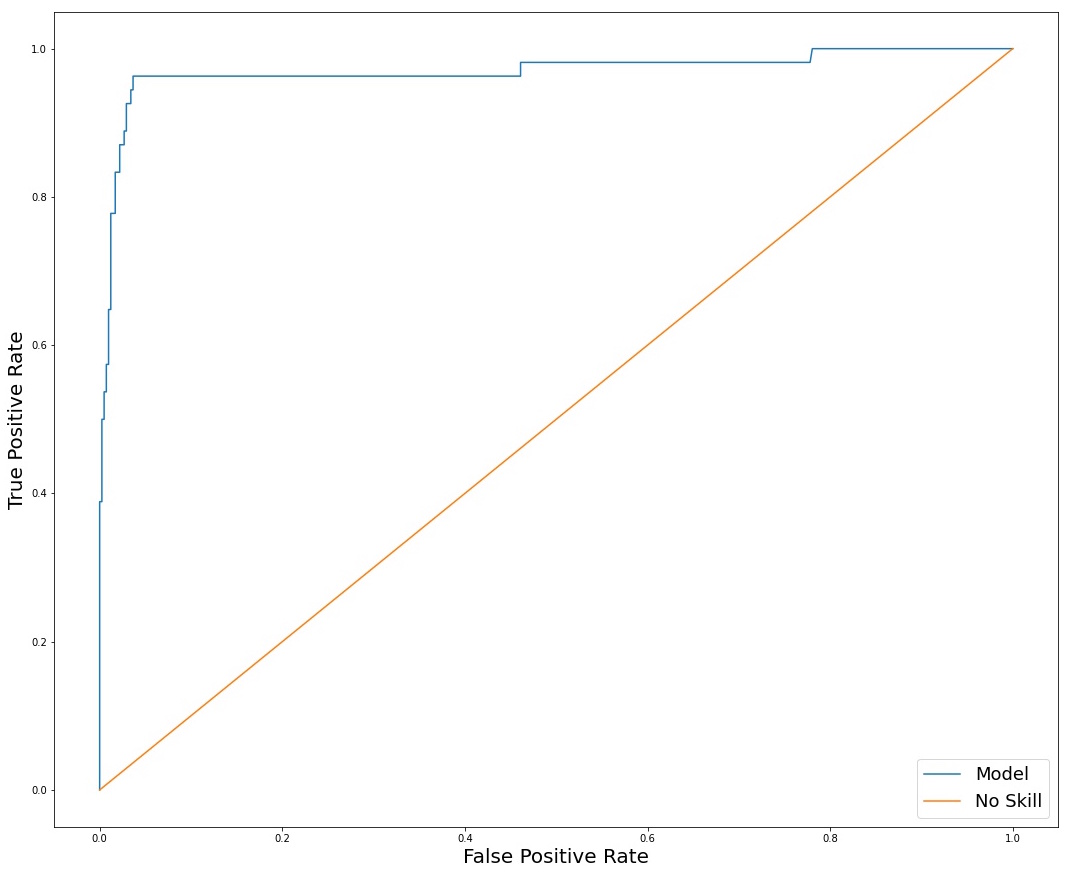}&
\includegraphics[width=0.33\textwidth]{ 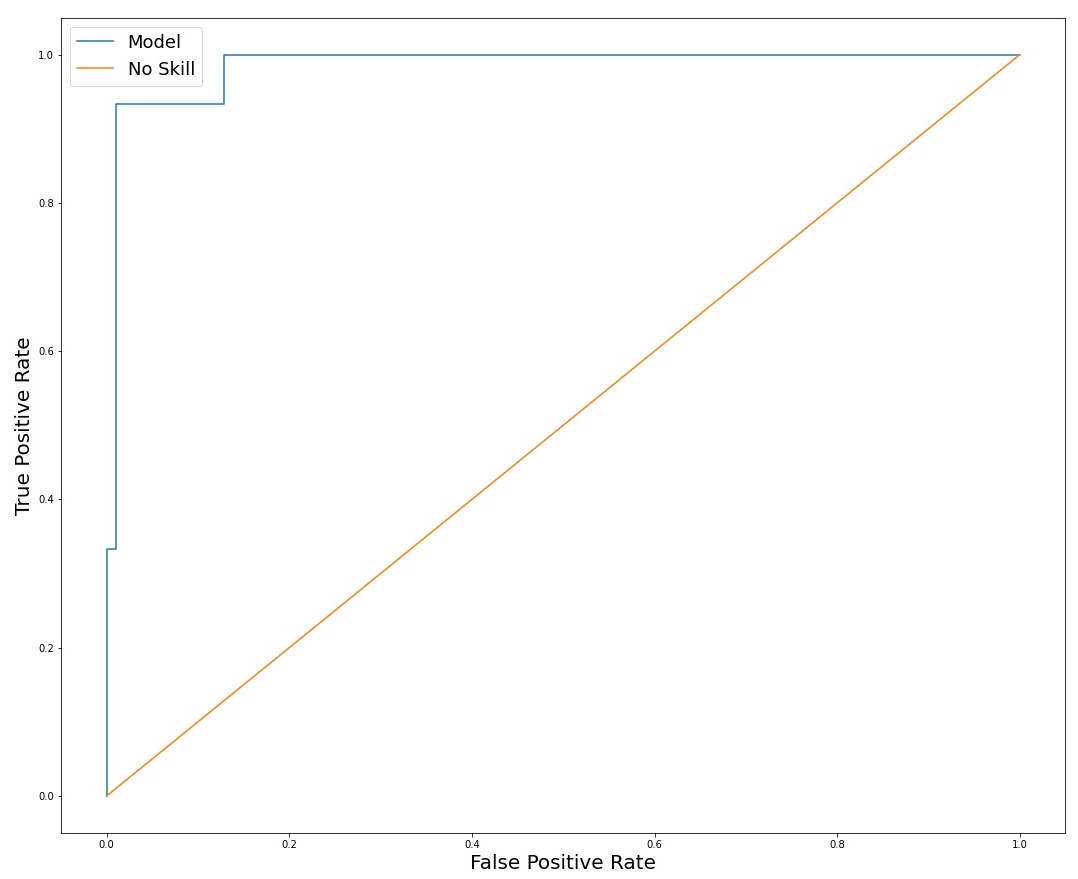}&
\includegraphics[width=0.33\textwidth]{ 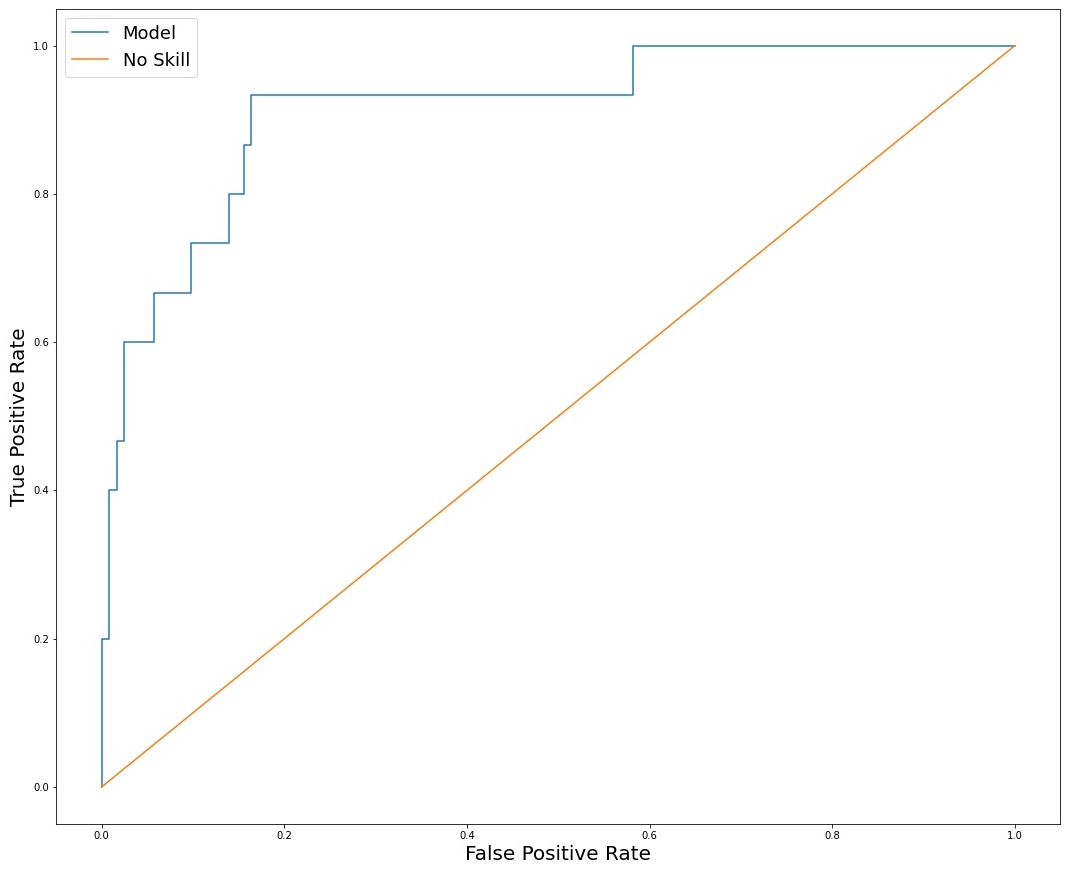}\\
(a)&(b)&(c)\\

\includegraphics[width=0.33\textwidth]{ 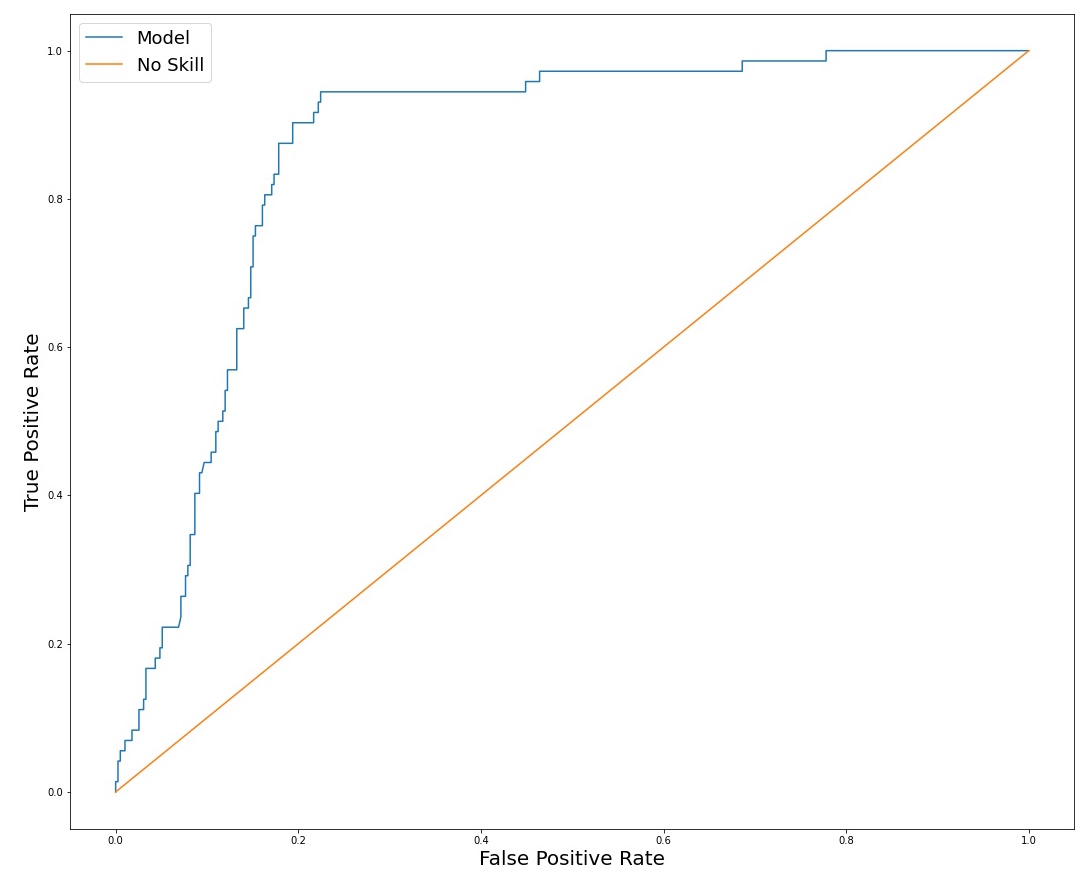}&
\includegraphics[width=0.33\textwidth]{ 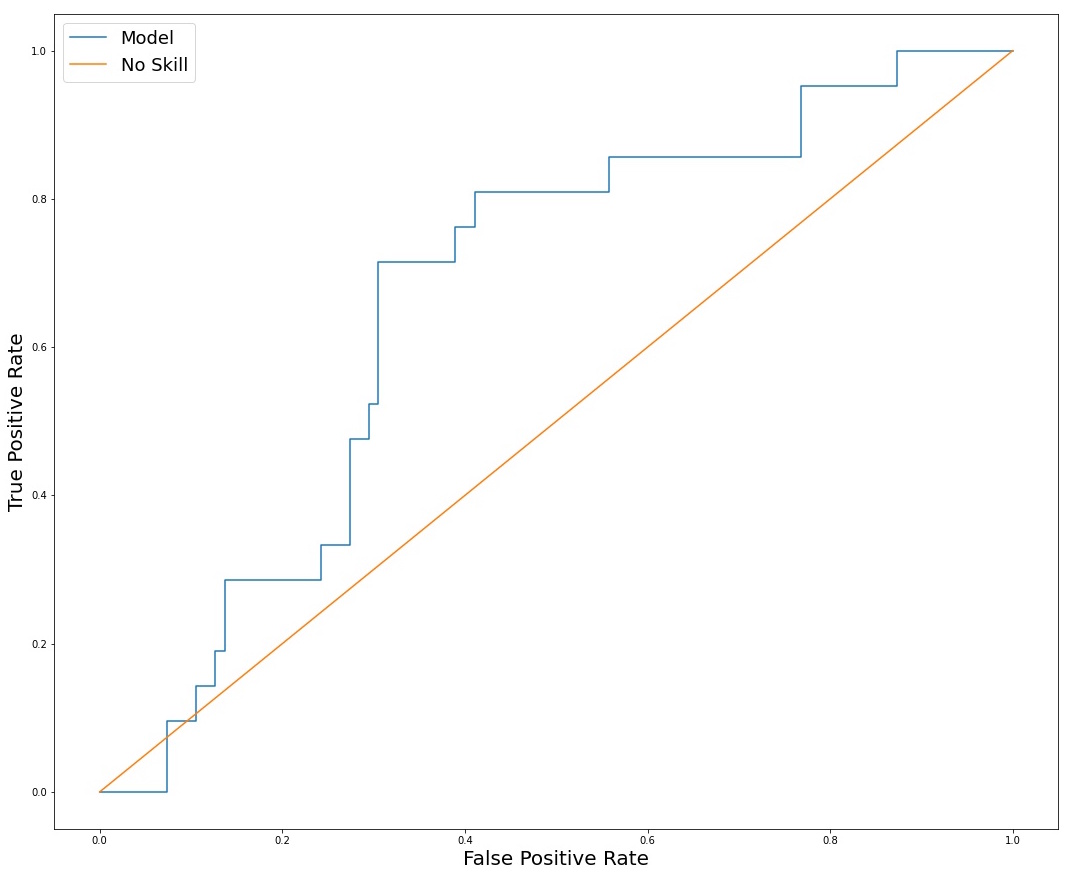}&
\includegraphics[width=0.33\textwidth]{ 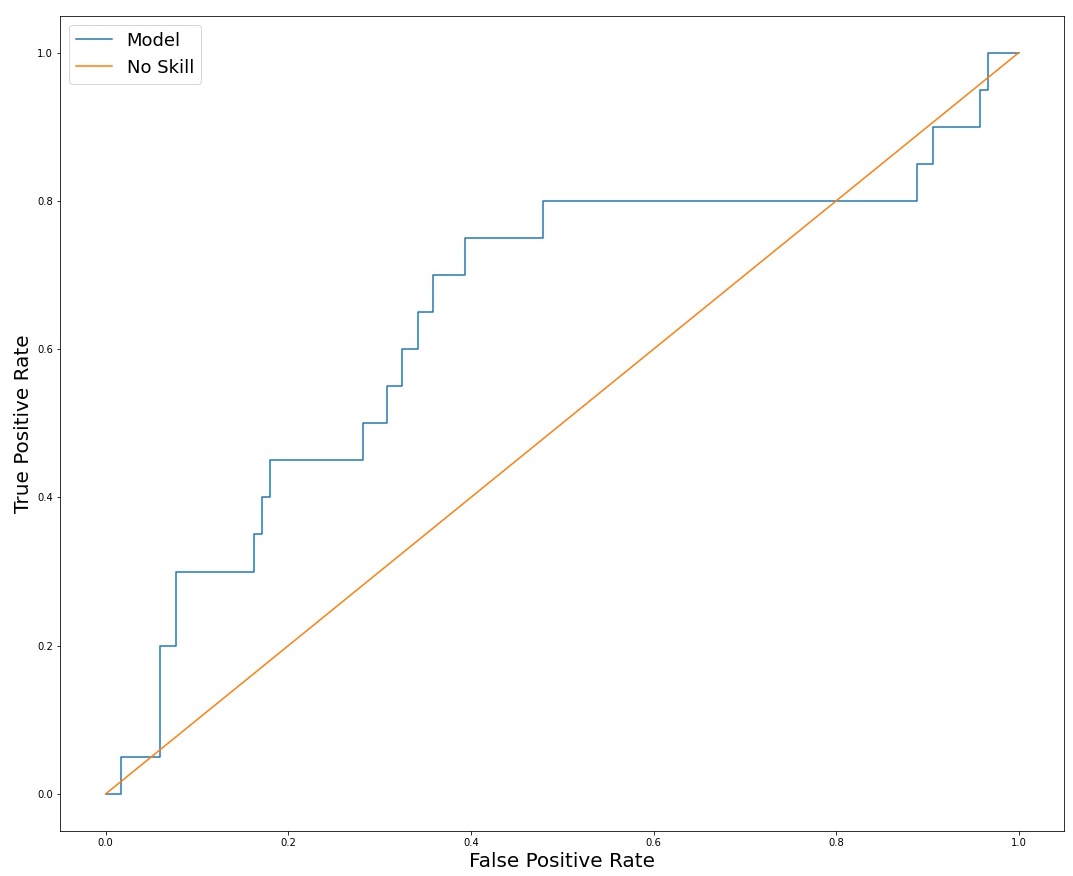}\\
(d)&(e)&(f)\\
    \end{tabular}
    \caption{Receiver Operating Characteristic Curves for line chart classification in (a) training set, (b) validation set, (c) test set and that of pie chart classification in (d) training set, (e) validation set and (f) test set.}
    \label{fig:ROCstage3}
\end{figure}

For pie chart recognition, we set the batch size to 128 and the learning rate to 4e-4. As we have a highly imbalanced dataset, we achieve good enough results in terms of MCC, scoring of 0.22 in the test set as shown in Table~\ref{s3_final}. The obtained auROC for pie charts is 0.64 in the test set. We avail a better result in terms of recall or sensitivity of 0.94 in the training set, 0.71 in the validation set and 0.75 in the test set. For line charts, we set the batch size to 256 and the learning rate remains as default to 1e-3. In Table \ref{s3_final}, we find outstanding results in terms of auROC score of 0.96 in the training set, 0.98 in the validation set and over 0.91 in the test set. Our obtained MCC in the train, validation and test sets is 0.96, 0.92 and 0.51 which is a better score than the prediction of pie charts. The ROC analysis for both of the tasks are given in Fig.~\ref{fig:ROCstage3}.

\begin{table}[t]
\caption{Experimental results for chart type prediction task.}
\label{s3_final}
\centering
\begin{tabular}{|p{1.5cm}|p{2.2cm}|c|c|c|c|}
\hline 
    \bf problem &\bf dataset &\bf Specificity &\bf Sensitivity&\bf MCC&\bf auROC\\ \hline
    \multicolumn{1}{|c|}{\multirow{3}{*}{Pie Chart}}&Training set&0.742&0.944&0.51&0.86\\
    \multicolumn{1}{|c|}{}&Validation set&0.6945&0.714&0.32&0.66\\ 
    \multicolumn{1}{|c|}{}&Test set&0.573&0.75&0.22&0.64\\ \hline
    \multicolumn{1}{|c|}{\multirow{3}{*}{Line Chart}}&Training set&0.9634&0.963&0.96&0.96\\
    \multicolumn{1}{|c|}{}&Validation set&0.990&0.933&0.92&0.98\\ 
    \multicolumn{1}{|c|}{}&Test set&0.893&0.733&0.51&0.91\\ \hline
\end{tabular}
\end{table}

\begin{figure}[!htb]
    \centering
      \includegraphics[width=0.8\linewidth]{ 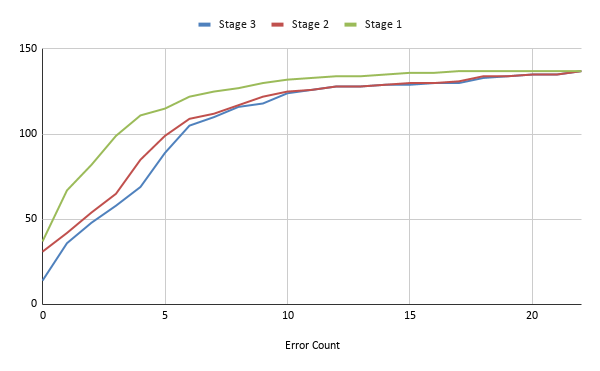}
      \caption{Cumulative frequency of error of three states put in a pipeline on the test set.}
      \label{error_Cu_by}
    \end{figure}
\subsection{Overall Performance}

In order to discuss the overall performance of our work, we have created a pipeline same as shown in Fig.~\ref{archi_1}. Our pipeline merges all the stages of our work and outputs the results we have already discussed and shown in this section. After obtaining the final results, we have checked for all possible errors occurs after completion of each stages. After completing stage 1, if both of the entity set have similar number of entities $(N = M)$ then we consider 1-to-1 sequential mapping. The cumulative frequency of error count for each of the stages is shown in Fig.~\ref{error_Cu_by}. This plot shows how each stage cumulatively produces error in the pipeline. However, we notice that although we have a good number of samples without error, there are room to improve and as shown in the figure, the most error-prone task is the task 3 due to the poor performance in pie chart type prediction. We also show one partially correct and one fully correct chart examples generated by {\methodname} in Table~\ref{tab:finalRes}.

\begin{table}[!htb]
    \caption{Sample input and outputs of {\methodname}.}
    
    \centering
    
\begin{footnotesize}
\begin{tabular}{|c|c|p{10cm}|}
\hline 
 \bf Input &\bf Sample text &Tzuyu is a gaming expert . She surveyed 200 individuals to judge the popularity of the video games among her all time favorites . After her survey she concluded that 25 people voted for World of Warcraft , 46 voted for Black Ops , 12 voted for Overwatch , 25 for Modern Warfare , 30 for PUBG , 50 for Sims and 40 for Assassin ' s Creed . \\
\hline
\bf Output &\bf $x$ entities&['World of Warcraft', 'Black Ops', 'Overwatch', 'Modern Warfare', 'PUBG', 'Sims', 'Assassin', 's Creed']\\
 &\bf $y$ entities&['25', '12', '25', '30', '50', '50', '40', '40']\\
 &\bf chart type&['bar']\\
 \hline 
 \hline 
 
 \bf Input &\bf Sample text& Mr . Jamal worked in the Meteorological Department for 8 years . He noticed a strange thing in recent times . On certain days of the month , the weather varied strongly . He wrote down the information to make a pattern of the event . The information of the paper is as follows : on the 3rd day of the month the temperature is 36 degrees Celsius , 7th day is 45 degrees Celsius , 9th day is 18 degrees Celsius , 11th day is 21 degrees Celsius , 17th day is 9 degrees Celsius , 19th day is 45 degrees Celsius , 21st day is 36 degrees Celsius , 27th day is 21 degrees Celsius and 29th day is 45 degrees Celsius . He finds a weird pattern in these dates and makes a report and sends it to his senior officer .\\
 \hline 
\bf  Output &\bf $x$ entities&['3rd day', '7th day', '9th day', '11th day', '17th day', '19th day', '21st day', '27th day', '29th day']\\
 &\bf $y$ entities&['36', '45', '18', '21', '9', '45', '36', '21', '45']\\
 &\bf chart type&['bar', 'Line']\\
 \hline 

    \end{tabular}
\end{footnotesize}
\label{tab:finalRes}
\end{table}

\section{Conclusion \label{secCon}}

In this paper we have presented {\methodname}, an automatic multi-staged technique that is able to generate charts from human written analytical text. Our technique has been tested on a dataset curated for this task. Despite having a short corpora, {\methodname} provides satisfactory results in every stage regarding automatic chart generation. One of the limitation of our work is the size of the dataset. With a larger dataset, we believe the methodology presented in this paper will provide further improved results. {\methodname} is currently limited to the prediction of only three basic chart types: bar charts, pie charts and line charts. It is possible to extend it for further types. Recently a dataset for chart-to-text has been proposed in \cite{obeid2020chart}. It is possible to use that dataset for the reverse problem also. We believe, it is possible to tune and experiment with more types of suitable neural architecture further for all the stages to improve overall accuracy.

\bibliographystyle{unsrt}
\bibliography{ref}
\end{document}